\documentclass[lettersize,journal]{IEEEtran}
\usepackage{amsmath,amsfonts}
\usepackage{algorithmic}
\usepackage{algorithm}
\usepackage{array}
\usepackage[caption=false,font=normalsize,labelfont=sf,textfont=sf]{subfig}
\usepackage{textcomp}
\usepackage{stfloats}
\usepackage{url}
\usepackage{verbatim}
\usepackage{graphicx}
\usepackage{multirow}
\usepackage{rotating} 
\usepackage{textcomp} 
\usepackage{cite}
\usepackage{booktabs} 
\usepackage[dvipsnames,svgnames,table]{xcolor} 
\def\BibTeX{{\rm B\kern-.05em{\sc i\kern-.025em b}\kern-.08em
    T\kern-.1667em\lower.7ex\hbox{E}\kern-.125emX}}
\begin{document}

\title{RmGPT: A Foundation Model with Generative Pre-trained Transformer for Fault Diagnosis and Prognosis in Rotating Machinery}

\author{
    Yilin Wang,
    Yifei Yu,
    Kong Sun,
    Peixuan Lei,
    Yuxuan Zhang,
    Enrico Zio,~\IEEEmembership{Fellow,~IEEE,}
    Aiguo Xia,
    and Yuanxiang Li,
    \thanks{Yilin Wang is with the School of Aeronautics and Astronautics, Shanghai Jiao Tong University, Shanghai 200240, China, and also with the Shanghai Innovation Institute, Shanghai, China (e-mail: pandalin@sjtu.edu.cn).}%
    \thanks{Yifei Yu, Kong Sun, Peixuan Lei, Yuxuan Zhang, and Yuanxiang Li are with the School of Aeronautics and Astronautics, Shanghai Jiao Tong University, Shanghai 200240, China (e-mails: \{yuyifei, mr\_sun, lei.2333, yuxuanzhang, yuanxli\}@sjtu.edu.cn).}%
    \thanks{Enrico Zio is with MINES Paris-PSL University, Paris, France, and the Energy Department, Politecnico di Milano, Milano, Italy (e-mail: enrico.zio@polimi.it).}%
    \thanks{Aiguo Xia is with the Beijing Aeronautical Technology Research Center, Beijing, China (e-mail: xag14@tsinghua.org.cn).}%
    \thanks{Corresponding author: Yuanxiang Li (e-mail: yuanxli@sjtu.edu.cn).}%
}

\markboth{Journal of \LaTeX\ Class Files,~Vol.~14, No.~8, August~2021}%
{Shell \MakeLowercase{\textit{et al.}}: A Sample Article Using IEEEtran.cls for IEEE Journals}


\maketitle

\begin{abstract}
In industry, the reliability of rotating machinery is critical for production efficiency and safety. Current methods of Prognostics and Health Management (PHM) often rely on task-specific models, which face significant challenges in handling diverse datasets with varying signal characteristics, fault modes and operating conditions. Inspired by advancements in generative pretrained models, we propose RmGPT, a unified model for diagnosis and prognosis tasks. RmGPT introduces a novel generative token-based framework, incorporating Signal Tokens, Prompt Tokens, Time-Frequency Task Tokens and Fault Tokens to handle heterogeneous data within a unified model architecture. We leverage self-supervised learning for robust feature extraction and introduce a next signal token prediction pretraining strategy, alongside efficient prompt learning for task-specific adaptation. Extensive experiments demonstrate that RmGPT significantly outperforms state-of-the-art algorithms, achieving near-perfect accuracy in diagnosis tasks and exceptionally low errors in prognosis tasks. Notably, RmGPT excels in few-shot learning scenarios, achieving 82\% accuracy in 16-class one-shot experiments, highlighting its adaptability and robustness. This work establishes RmGPT as a powerful PHM foundation model for rotating machinery, advancing the scalability and generalizability of PHM solutions. \textbf{Code is available at:} \url{https://github.com/Pandalin98/RmGPT}.
\end{abstract}



\section{Introduction}
Rotating machinery is a critical component in many industrial applications\cite{fink_potential_2020,zio_prognostics_2022,Sketching}, and ensuring its reliability is essential for production  efficiency and safety\cite{liu_artificial_2018,Satellite,SAH-NET,wang2025data}. Prognostics and Health Management (PHM) encompasses a range of methodologies aimed at assessing and predicting the present and future health status of equipment\cite{zhang2023realistic,wang2024physics,Matrix}. The outcomes of PHM enable timely maintenance to prevent unexpected failures. Current PHM methods for fault diagnosis and prognosis often rely on task-specific models tailored to particular types of equipment and operational conditions and specific fault modes. They typically involve handcrafted features and machine learning algorithms designed to identify specific fault patterns from historical data\cite{gao_data_2024,wang_establishment_2022,wen_gru-ae-wiener_2024,Microgrid}.

While effective in specific contexts, these task-specific models encounter significant challenges when applied to diverse datasets with varying signal characteristics, fault modes, and operating environments\cite{wang2023self,BasGPT}. Several key difficulties arise in this context: \textbf{1) Variability in Signal Dynamics:} Different types of diagnosis equipment exhibit significant variations in sensor quantity, installation position, monitored types of signals, and sampling frequencies. Modeling these diverse signals within a unified framework is highly challenging. \textbf{2) Diversity in Fault Mechanisms and Patterns:} The wide range of equipment types and their differing designs and fault mechanisms complicate the definition of unified diagnosis and prognosis tasks. Each equipment type may exhibit unique fault behaviors that are difficult to generalize across other types. \textbf{3) Lack of a Foundation Model for Rotating Machinery:} The differences in input signals and diagnostic tasks lead to the difficulty of having a foundation model capable of using a single set of parameters to effectively perform diagnosis and prognosis tasks across multiple types of equipment. Despite efforts by researchers to employ transfer learning techniques for domain adaptation and generalization in rotating machinery\cite{10473026,10577994,10418185,9951150,sun2023unsupervised,10225508}, the absence of such a foundation model limits knowledge transfer and sharing between different devices and hinders the application of AI techniques in real PHM scenarios\cite{li2023chatgpt}.

Recent advancements in generative pre-trained models, such as ChatGPT \cite{zhou2023chatgpt} and SAM \cite{kirillov2023segment}, have demonstrated the power of self-supervised learning in capturing generalizable patterns from vast amounts of unlabeled data. Inspired by these breakthroughs, we propose \textbf{RmGPT}, a generative foundation model designed to adapt to diverse fault diagnosis and prognosis tasks in rotating machinery. Unlike traditional discriminative models that rely on manually crafted features or task-specific finetuning, RmGPT takes a generative approach, modeling the machinery's health status by learning to predict and generate the signal patterns from raw data.

\begin{figure}[!htbp]
\centerline{\includegraphics[width=1\columnwidth]{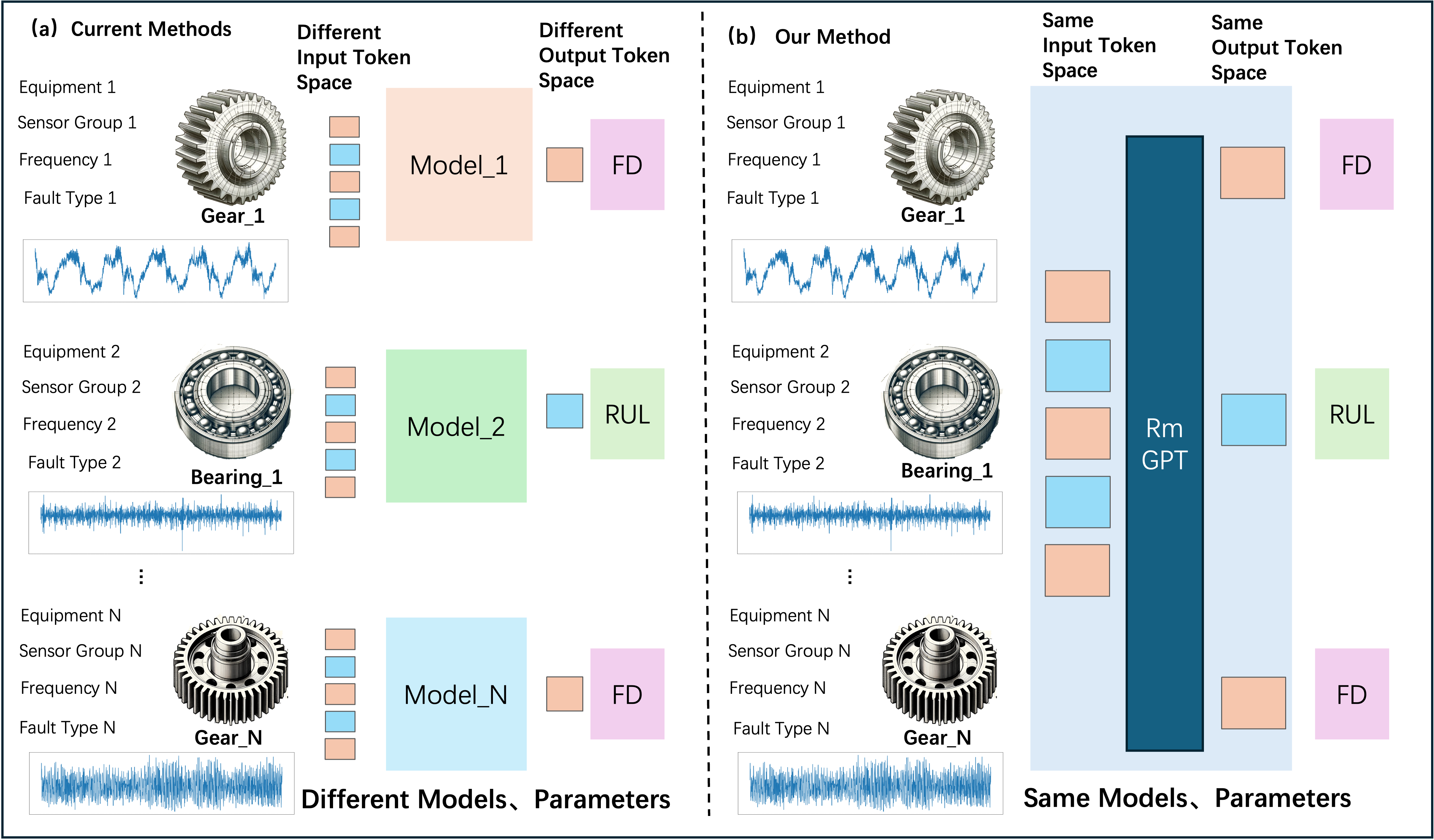}}
\caption{Unlike current research methods that use different models for different equipment, this study uses the same model parameters to address cross-condition diagnosis and prognosis for rotating machinery with varying sensors, frequencies, and fault modes.}
\label{fig:intro_pic}
\end{figure}

RmGPT introduces a unified token-based framework that incorporates \textit{Signal Tokens}, \textit{Prompt Tokens}, \textit{Time-Frequency Task Tokens}, and \textit{Fault Tokens}. This design enables the model to autonomously adapt to varying signal characteristics and fault modes across different equipment types. By learning to generate relevant health information through these tokens, RmGPT produces a robust representation of machinery health. {Although inspired by generative transformer architectures, RmGPT is not a language model. It is specifically designed for multivariate time-series signals and operates entirely on continuous sensor data without using textual inputs or natural language supervision.} Leveraging self-supervised learning and next-token prediction, the model efficiently captures the underlying structures in diverse signal data, ensuring high generalization across datasets with minimal labeled data. As shown in Fig.~\ref{fig:intro_pic}, this approach empowers RmGPT to generate consistent health representations that are seamlessly integrated into downstream tasks, making it highly adaptable and capable of handling different equipment and operational conditions.

In this paper, we present several key contributions:

\textbf{1. Unified Diagnosis and Prognosis Paradigm:} We propose a novel generative paradigm based on a semantic token space for diagnosis and prognosis tasks. This paradigm leverages learnable \textit{Fault Tokens} to capture fault prototypes from different equipment types and \textit{Time-Frequency Task Tokens} to represent the health status semantics of input signals. By comparing these generative health status semantics tokens with Fault Tokens, RmGPT performs unified diagnosis and prognosis across varying equipment and fault conditions.

\textbf{2. Unified Efficient Model for Adaptive Signal Processing:} We introduce an efficient generative model that can adaptively handle varying signal inputs from different equipment types. The model utilizes learnable \textit{Prompt Tokens} to capture sensor-specific semantics and employs an efficient \textit{Signal Token} mapping method. A dual-stage attention mechanism is used to generatively extract and refine signal features. Additionally, we propose a \textit{Next Signal Token Prediction} pretraining strategy, which enables RmGPT to capture generalizable features from vast amounts of unlabeled data, and an \textit{Efficient Prompt Learning} method for rapid adaptation to new tasks.

\textbf{3. Unified Superior Performance Across Diverse Datasets:} Extensive experiments demonstrate that RmGPT outperforms state-of-the-art (SOTA) algorithms in both diagnosis and prognosis tasks. The model achieves near-perfect accuracy in diagnosis tasks and exceptionally low errors in prognosis tasks. Moreover, RmGPT excels in few-shot learning scenarios, maintaining high accuracy even with minimal training samples, thus validating its generative robustness and adaptability across diverse datasets and equipment types.

The remainder of this paper is structured as follows. Section II describes the problem definition and the proposed RmGPT model. Section III details the training process, including pretraining and prompt learning. Section IV presents the experimental setup and results, highlighting the model's performance across various tasks. Finally, Section V concludes the paper with a summary of our findings and directions for future research.

\section{Problem Definition}
We consider  PHM tasks of diagnosis and prognosis for rotating machinery. Different equipment generate datasets $D=\left\{D_i \mid i=1, \ldots, n\right\}$, each described as $D_i=\left(\mathcal{X}_i, \mathcal{Y}_i\right)$, where $\mathcal{X}_i$ and $\mathcal{Y}_i$ represent the time series of the input signals and the corresponding task outputs, respectively. Each $\mathcal{X}_i$ consists of sensor monitored signals with varying dimensions $M_i$, different frequencies $f_i$, distinct operational conditions, and various task types $\mathcal{Y}_i$. The datasets $D_i$, thus, exhibit significant heterogeneity. Traditional methods have individually defined the inputs and outputs of the problem for specific operational conditions and fault types of a particular device or class of equipment. This Section attempts to offer a description of the diagnosis and prognosis tasks of different $D_i$ in a unified form.

\subsection{Definition and Characteristics of Foundation Models in PHM}

{
Foundation models have recently emerged as a dominant paradigm in natural language processing and computer vision, where a single, pre-trained model can be adapted to various downstream tasks with minimal task-specific adjustment. However, applying this concept directly to Prognostics and Health Management (PHM)—particularly in the context of rotating machinery—requires careful reinterpretation, due to the structured, heterogeneous, and domain-specific nature of industrial data.
}

{
In this work, we define a \textbf{PHM foundation model} as:
\begin{quote}
    \textit{A general-purpose model $F$ with shared parameters $\theta$, trained via self-supervised learning on heterogeneous time-series datasets $\{D_i\}$, capable of producing task-relevant representations $\mathbf{h}_i = F(\mathcal{X}_i; \theta)$ and supporting diverse diagnostic and prognostic tasks across different equipment and conditions without modifying the model architecture.}
\end{quote}
}

{
Compared to traditional approaches where each dataset $D_i = (\mathcal{X}_i, \mathcal{Y}_i)$ is handled by an independent model $F_i$ with task-specific parameters $\theta_i$, a PHM foundation model aims to unify representation learning and task modeling through a consistent parameter set $\theta$ shared across all datasets.
}

{
We summarize the key properties of a PHM foundation model as follows:}
\begin{itemize}
    \item {\textbf{Cross-domain representation learning:} The model is trained on a diverse collection of datasets $\{D_i\}$ with varying input dimensions, sampling frequencies, fault types, and operating conditions, and is expected to learn generalizable representations $\mathbf{h}_i$ from unlabeled data.}

    \item {\textbf{Multi-task flexibility:} The learned representations should support a range of PHM tasks (e.g., Fault Diagnosis, Remaining Useful Life (RUL) prediction) in a consistent framework without requiring redesign of task-specific modules.}

    \item {\textbf{Architecture-level unification:} The model maintains a consistent network architecture across datasets and tasks, facilitating adaptation to new application scenarios without structural modification.}
\end{itemize}

{
This reformulation of the foundation model paradigm in PHM provides a theoretical basis for constructing general-purpose models that unify diagnosis and prognosis across rotating machinery datasets, while offering scalability to broader industrial applications.
}

\subsection{Self-Supervised Learning for Foundation Models}
Different  from a model $\mathcal{Y}_i=F_i(\mathcal{X}_i;\theta_i)$ and its parameters $\theta_i$ for specific data $D_i$, we aim to find a foundation model $Y=F(\mathcal{X};\theta)$ for all rotating machinery datasets $D$. A feasible approach is to leverage the powerful capabilities of self-supervised learning, which excels in learning robust representations from unlabeled data and unifying diverse tasks and conditions.

In self-supervised learning, a pretext task $T_p$ is designed to train the model using unlabeled data $\mathcal{X}$. The objective is to learn a general representation $\mathbf{h} = F(\mathcal{X};\theta_e)$, where $\theta_e$ are the pretraining parameters. The representation $\mathbf{h}$ is then used to solve the pretext task $T_p$, typically by predicting some part of the input $\mathcal{X}$ based on the rest. Formally, the self-supervised learning objective is:
\begin{equation}
\theta_e = \arg \min_{\theta_e} \mathbb{E}_{\mathcal{X} \sim D} \left[ \mathcal{L}_{T_p}(\mathcal{X}, F(\mathcal{X};\theta_e)) \right],
\end{equation}
where $\mathcal{L}_{T_p}$ is the loss function associated with the pretext task $T_p$. Once pre-trained, the model $F$ and representation $\mathbf{h}$ can be fine-tuned on downstream diagnosis and prognosis tasks using labeled data $\mathcal{Y}$. This pretraining on extensive unlabeled data enables the model to learn generalized features, making it adaptable for various diagnosis and prognosis tasks, thus providing a comprehensive and flexible framework for rotating machinery PHM.

\subsubsection{Diagnosis and Prognosis}
Building on the general representation $\mathbf{h}$ obtained through pretraining, we define the mapping relationship for task outputs as:
\begin{equation}
\hat{\mathcal{Y}}_i = \arg \max_{\mathcal{Y}_i} P(\mathcal{Y}_i \mid \mathbf{h}_i, \theta_d),
\end{equation}
where $\hat{\mathcal{Y}}_i$ is the predicted output for dataset $D_i$ based on the learned tokens $\mathbf{h}_i$ and the same foundation model $F$ with the same parameters $\theta_d$. For the diagnosis task, ${Y}_i \in \mathbb{R}^{FC_i}$, the model outputs a classification among the fault categories $FC_i$. For the prognosis task, ${Y}_i \in \mathbb{R}$, the model outputs a RUL prediction.

\subsubsection{Prompt Learning\& Finetuning}
Prompt learning and finetuning are two approaches to adapt the foundation model to downstream tasks, with different optimization parameters. 

For prompt learning, task-specific prompts $\mathbf{p}$ are introduced to guide the model's adaptation:
\begin{equation}
\theta_p = \arg \min_{\theta_p} \mathbb{E}_{(\mathcal{X}_i, \mathcal{Y}_i) \sim D} \left[ \mathcal{L}(\mathcal{Y}_i, F(\mathcal{X}_i, \mathbf{p}; \theta_e)) \right],
\end{equation}
where $\theta_e$ are the pretrained parameters and $\theta_p$ are the parameters for prompt adaptation. The prompts $\mathbf{p}$ help the model leverage the pretrained representations effectively for specific tasks.

For finetuning, the model parameters $\theta_f$ are directly adjusted based on the labeled data:
\begin{equation}
\theta_f = \arg \min_{\theta_f} \mathbb{E}_{(\mathcal{X}_i, \mathcal{Y}_i) \sim D} \left[ \mathcal{L}(\mathcal{Y}_i, F(\mathcal{X}_i; \theta_f)) \right],
\end{equation}
notably the finetuned parameters $\theta_f$ and  $\theta_p$ are the same for all the data and tasks.

\subsubsection{Few-shot Learning \& One-Shot Learning}
Few-shot learning enables the model to perform well with limited labeled data. The objective is defined as:
\begin{equation}
\theta_{fs} = \arg \min_{\theta_{fs}} \mathbb{E}_{(\mathcal{X}_i, \mathcal{Y}_i) \sim D_{fs}} \left[ \mathcal{L}(\mathcal{Y}_i, F(\mathcal{X}_i; \theta_e, \theta_{fs})) \right],
\end{equation}
where $\theta_{fs}$ are the parameters optimized using few-shot examples from $D_{fs}$, a subset of $D$ with few labeled samples. In particular, when each fault category in $D_{fs}$ has only one sample, the task becomes one-shot learning.

\section{Rotating Machinery Generative Pretrained Transformer}
\begin{figure*}[hbtp]
    \centering
    \includegraphics[width = 0.9\textwidth]{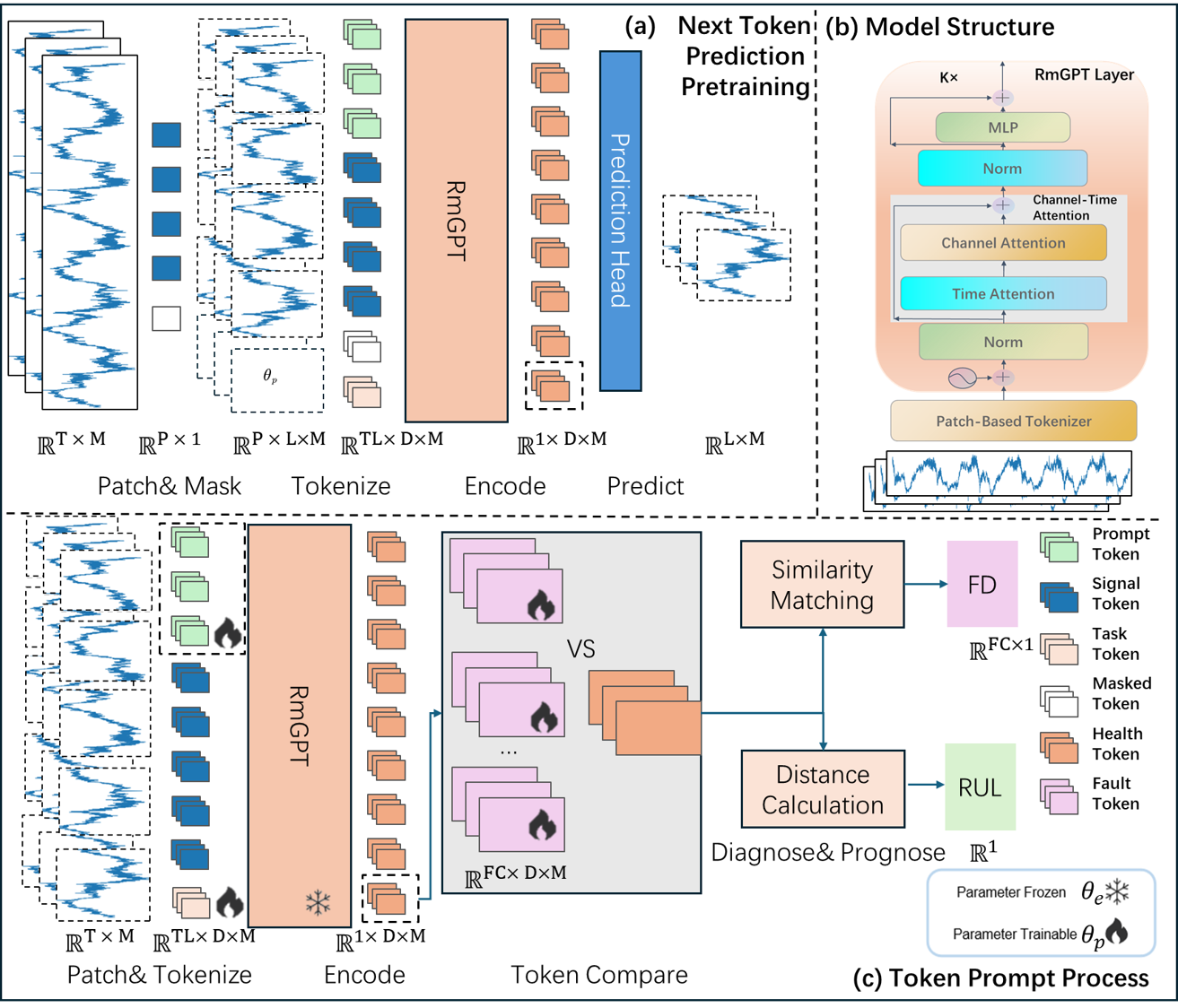}
    \caption{The proposed foundation model for diagnosis and prognosis, RmGPT, includes: (a) a pretraining strategy for predicting future signal tokens, (b) the internal architecture of RmGPT, and (c) a generalized paradigm for diagnosis and prognosis based on generative tokens and the prompt technique.}
    \label{fig:framework}
\end{figure*}

In this chapter, we introduce RmGPT, as depicted in Fig.~\ref{fig:framework}. RmGPT serves as a foundation model for unified diagnosis and prognosis of rotating machinery. {Although RmGPT draws structural inspiration from large generative models, it is not a language model. Instead, it is specifically designed to process multivariate time-series signals in industrial scenarios and does not rely on textual inputs or language modeling objectives.} The model employs a generative framework using tokens to represent diagnosis and prognosis tasks. It features an optimized architecture with a patch-based tokenizer and a channel-time attention transformer, ensuring efficient processing and adaptability. The training process includes a pretrained strategy focused on predicting signal tokens and a prompt-based technique for finetuning the model to specific tasks.

\subsection{Unified Diagnosis and Prognosis Framework}

To develop a foundation model that unifies the modeling of $\mathcal{X}$ and the diagnosis and prognosis $\mathcal{Y}$, we draw inspiration from large language models and multimodal frameworks, proposing the use of tokens to represent diverse data and task information. Unlike conventional approaches that require separate models for different datasets or tasks, RmGPT employs a unified architecture with a consistent set of parameters to model signals $\mathcal{X}_i$ from various equipment, sensor groups, frequencies, and operational conditions, alongside their corresponding diagnosis and prognosis outputs $\mathcal{Y}_i$ for different fault categories. The model generates outputs based on health status semantic tokens, all within a cohesive token space.

We integrate the diagnosis and prognosis tasks of different equipment into a unified token sequence, where each token in the sequence is designed to represent specific types of information, as depicted in Fig.\ref{fig:tokens and attention}a. At the beginning of the sequence, Prompt Tokens are introduced to enhance the model's adaptability by embedding task-specific statistical information, enabling seamless integration of context-aware knowledge. The middle of the sequence is composed of Signal Tokens, which capture the essential trends and patterns within the input signals, ensuring consistency and stability across varying scales and sensor configurations. Towards the end of the sequence, Time-Frequency Task Tokens are designed to encapsulate the health semantics of the input data, providing a strong foundation for both diagnosis and prognosis tasks. Finally, Fault Tokens represent the prototype characteristics of different fault modes, facilitating precise fault detection and Remaining RUL prediction through a comparison-based approach. This structured token sequence allows for the execution of various downstream tasks based on the relationships between these tokens.

\subsubsection{Prompt Tokens}
Prompt tokens $T_p \in \mathbb{R}^{l_p \times M_i \times d}$ are crafted as a combination of sequence time-domain statistical information and learnable embeddings, where $l_p$ denotes the length of the prompt token sequence, $M_i$ represents the number of signal channels for different equipment, and $d$ corresponds to the number of hidden units in the model. This dual-purpose design allows the model to effectively capture and utilize key statistical features of the input signals. Specifically, the extracted signal mean and variance are integrated into the token space, thereby enhancing the model's ability to perceive and process statistical nuances while maintaining a focus on the morphological characteristics of the signals. Furthermore, in scenarios involving multiple equipment and tasks, each dataset and task is assigned its own set of learnable prompt tokens. These tokens are tailored to incorporate context-specific knowledge, thus helping the model distinguish between different equipment conditions and tasks. This design ensures that the model remains adaptable and context-aware, providing a robust foundation for accurate diagnosis and prognosis across a wide range of scenarios.

\subsubsection{Signal Tokens}
To ensure that the signal tokens $T_s \in \mathbb{R}^{l_s \times M_i \times d}$ effectively capture the trends within the signals and maintain stability across varying scales, we first standardize the input signals by normalizing their mean and variance, where $l_s$ represents the length of the signal token sequence. This standardization step helps to stabilize the numerical values of the signals, ensuring consistency and reliability in the token representation. Furthermore, to provide more robust and coherent semantics for each token block, we segment the long time series signals into several sub-sequences, mapping each sub-sequence into a fixed-dimensional token space. Given that different equipment may have varying numbers of sensor channels $M_i$, we preserve the channel dimension's shape, allowing the model to process data from different channel configurations within a unified framework. Additionally, we propose a flexible network structure that can dynamically adapt to these varying channel numbers, further enhancing the model's ability to generalize across diverse datasets and equipment types.

\subsubsection{Time-Frequency Task Tokens \& Health Token}
Time-Frequency task tokens $T_t \in \mathbb{R}^{l_t \times M_i \times d}$ are designed for general representation learning across various tasks and can adapt to any diagnosis and prognosis tasks, where $l_t$ represents the length of the task token sequence. These tokens combine learnable embeddings and frequency-domain representations of the input signals. The input signals are transformed using the Fast Fourier Transform (FFT) to obtain the signal's spectrum and phase, which are then projected into the token space to get frequency-domain representations. A learnable embedding is concatenated before the frequency-domain representation to form the Time-Frequency task tokens. These tokens, containing task characteristics, are fed into the foundation model to produce Health Tokens $T_h \in \mathbb{R}^{l_t \times M_i \times d}$, representing the health state of the input signals for downstream tasks (Fig. \ref{fig:framework}c).

\begin{figure}[!htbp]
\centerline{\includegraphics[width=0.98\columnwidth]{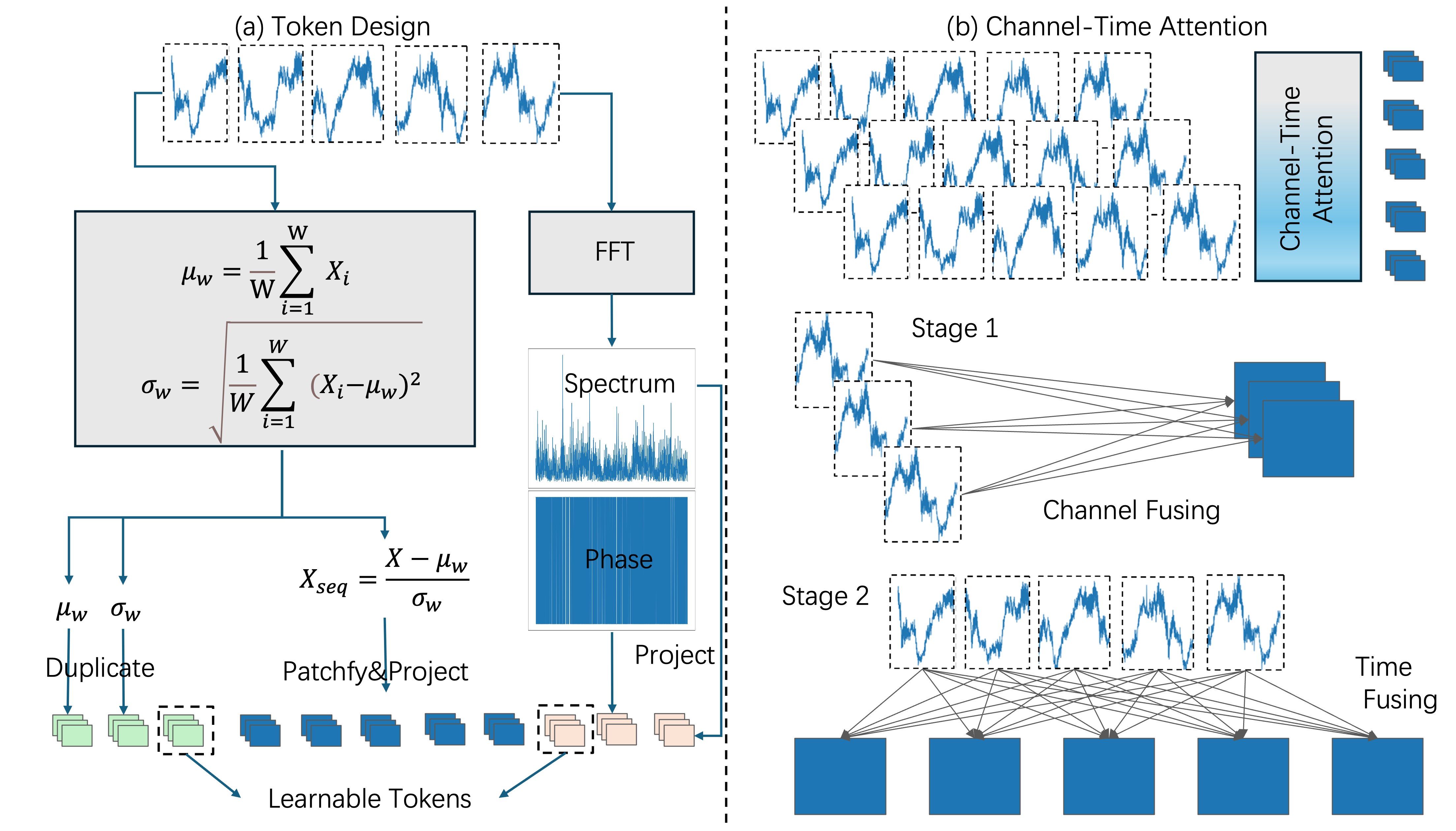}}
\caption{(a) Unified token design in the model's token space. (b) Channel-Time Attention mechanism for efficient aggregation of variable-length and multi-channel signals.}
\label{fig:tokens and attention}
\end{figure}

\subsubsection{Fault Tokens}
Fault tokens $T_f \in \mathbb{R}^{FC_i \times l_t \times M_i \times d}$ are introduced to address the challenges of generalizing fault diagnosis and prognosis across different equipment types and operational conditions. Here, $FC_i$ represents the fault types of the $i^{th}$ equipment, where a vector of shape $\mathbb{R}^{l_t \times M_i \times d}$ is stored for each fault type, encapsulating the characteristics of each channel under the respective fault condition. These tokens are designed as learnable embeddings that autonomously capture the prototype characteristics of various sensor channels under specific fault conditions during extensive data learning. The motivation behind using fault tokens lies in their ability to encapsulate the essential features of different fault modes, making them highly effective for tasks involving classification and prediction of RUL. By representing fault modes as prototype embeddings, the model can more accurately distinguish between different types of faults, even when the input signals vary significantly across equipment. This approach ensures that the diagnosis and prognosis tasks are grounded in the most relevant features of the fault, thereby enhancing the overall interpretability and robustness of the model.

{
To enhance the interpretability of the model's internal inference mechanism, as illustrated in Fig. 2(c), we provide a detailed explanation of how different types of tokens—particularly the Time-Frequency Task Tokens ($T_t$) and Fault Tokens ($T_f$)—interact during the inference process to support fault diagnosis and prognosis decisions.
}

{
Prior to the channel-time attention layers, we concatenate the Prompt Tokens ($T_p \in \mathbb{R}^{l_p \times M \times d}$), Signal Tokens ($T_s \in \mathbb{R}^{l_s \times M \times d}$), and Time-Frequency Task Tokens ($T_t \in \mathbb{R}^{l_t \times M \times d}$) along the sequence dimension to form the model's input:
\[
T_{\text{in}} = T_p \ \Vert \ T_s \ \Vert \ T_t \in \mathbb{R}^{N \times M \times d},
\]
where $N = l_p + l_s + l_t$ is the total sequence length. This unified token sequence is fed into the backbone encoder for joint modeling of channel and temporal dependencies:
\[
T_{\text{out}} = F_{\text{enc}}(T_{\text{in}}) \in \mathbb{R}^{N \times M \times d}.
\]
We then extract the output embeddings corresponding to the Time-Frequency Task Tokens from the last $l_t$ positions of $T_{\text{out}}$, and compute the Health Token $T_h \in \mathbb{R}^{M \times d}$ by applying mean pooling over the sequence dimension:
\[
T_h = \frac{1}{l_t} \sum_{i=N-l_t+1}^{N} T_{\text{out}}^{(i)},
\]
where $T_{\text{out}}^{(i)} \in \mathbb{R}^{M \times d}$ denotes the $i$-th token in the output sequence. This compact representation $T_h$ captures the aggregated temporal and frequency-domain health semantics of the system, and is subsequently compared with learned fault prototypes for diagnosis and prognosis.
}

{
Following the channel-time attention layers, the input signal is first converted into a series of Signal Tokens ($T_s \in \mathbb{R}^{l_s \times M \times d}$), where $l_s$ is the token length, $M$ is the number of sensor channels, and $d$ is the embedding dimension. These tokens are further processed and transformed into Time-Frequency Task Tokens ($T_t \in \mathbb{R}^{l_t \times M \times d}$), which capture both time-domain and frequency-domain characteristics of the input signals. The $T_t$ tokens are aggregated through attention mechanisms into Health Tokens ($T_h \in \mathbb{R}^{M \times d}$), serving as compact latent representations of the system's current health state.
}

{
During inference, $T_h$ is compared against a set of learnable Fault Tokens $T_f = \{T_f^{(1)}, T_f^{(2)}, \dots, T_f^{(C)}\}$, where each $T_f^{(c)} \in \mathbb{R}^{M \times d}$ represents the prototype embedding of fault class $c$, and $C$ is the total number of fault categories. For classification tasks (i.e., fault diagnosis), the model computes the similarity between $T_h$ and each $T_f^{(c)}$ using a vector similarity function $\text{Sim}(\cdot,\cdot)$ (e.g., cosine similarity). The predicted class label $\hat{y}_{\text{diag}}$ is assigned to the class with the highest similarity:
}
\begin{equation}
\hat{y}_{\text{diag}} = \arg\max_{c} \ \text{Sim}(T_h, T_f^{(c)})
\end{equation}

{
For regression tasks (i.e., Remaining Useful Life prediction), the model measures the Euclidean distance between $T_h$ and each $T_f^{(c)}$, and maps the smallest distance to an estimated RUL value $\hat{y}_{\text{rul}}$ via a transformation function $g(\cdot)$:
}
\begin{equation}
\hat{y}_{\text{rul}} = g\left(\min_{c} \|T_h - T_f^{(c)}\|\right)
\end{equation}

{
Here, $g(\cdot)$ is a task-specific regression function learned during training. This inference strategy enables the unified token representation to support both classification and regression tasks in a seamless and consistent manner.}

{
By comparing the semantic health state (captured by $T_t$ and aggregated into $T_h$) with class-level prototypes ($T_f$), the model achieves generalization across multiple fault types and signal domains. This prototype-driven matching mechanism, as depicted in Fig. 2(c), reflects the core design philosophy of RmGPT: a unified, interpretable, and extensible inference framework built upon token-based semantic alignment in a latent space.
}

\subsection{Model Architecture}
\label{sec:Model Structure}
The RmGPT model is meticulously designed to achieve efficient processing of input/output data while maintaining a high degree of flexibility and generality. The architecture incorporates several innovative components that contribute to its robust performance across various diagnostic and prognostic tasks, regardless of the equipment involved.

\subsubsection{Patch-Based Tokenizer}
{
The Patch-Based Tokenizer plays a crucial role in segmenting input signals into manageable patches \cite{wang_incorporating_2024}, transforming local information into semantic vectors suitable for processing by the Transformer network, as illustrated in Fig.~\ref{fig:tokens and attention}a. Given a multivariate time series $x^{(i)} \in \mathbb{R}^{L_i \times M_i}$, where $L_i$ is the sequence length and $M_i$ is the number of sensor channels, we first segment each univariate channel independently into $l_s^{(i)}$ temporal patches of fixed length $P$ and stride $S$, where:
\begin{equation}
l_s^{(i)} = \left\lfloor\frac{L_i - P}{S}\right\rfloor + 1
\end{equation}
This converts each 1D input sequence of length $L_i$ into a sequence of $l_s^{(i)}$ localized subsequences (patches), preserving the temporal continuity within each segment.

Each patch is then projected into a $d$-dimensional latent space using a shared linear transformation across channels, resulting in a compact embedding. By applying this process to all channels and stacking the results, we obtain a token tensor of shape:
\begin{equation}
x_t^{(i)} = \operatorname{Linear}(x_p^{(i)}) \in \mathbb{R}^{l_s^{(i)} \times M_i \times d}
\end{equation}
where the first dimension indexes temporal patches, the second dimension corresponds to sensor channels, and the third dimension is the latent embedding.

To encode temporal order, a positional encoding matrix $W_{pos} \in \mathbb{R}^{l_s^{(i)} \times d}$ is added to each channel’s token sequence via broadcasting:
\begin{equation}
T_s^{(i)} = x_t^{(i)} + W_{pos}
\end{equation}
The resulting matrix $T_s^{(i)} \in \mathbb{R}^{l_s^{(i)} \times M_i \times d}$ serves as the input to the Transformer encoder.

This tokenization strategy offers two key advantages. First, it significantly reduces the computational and memory cost of self-attention, which scales quadratically with sequence length \cite{nie2022time}. By converting the input from $L_i$ time points to approximately $L_i/S$ patch tokens, the model achieves substantial efficiency gains—for example, a sequence of length $L=2048$ with $P = S = 256$ results in only $l_s = 8$ tokens per channel. Second, representing each patch as a semantically meaningful unit allows the model to capture richer local temporal dynamics and structural patterns, leading to more robust and generalizable representations.
}

\subsubsection{Channel-Time Attention Transformer}

A key innovation in the RmGPT architecture is the Channel-Time Attention Transformer, which is designed to handle the variability in input data effectively while reducing computational complexity. This mechanism is motivated by the need to process multi-channel rotating machinery data efficiently.

The Channel-Time Attention Transformer operates through a two-stage attention mechanism, which is crucial for enabling the model to efficiently process signals from any combination of sensor channels. The first stage, Channel Attention, aggregates tokens from different channels within the same time block to extract the signal semantics:
\begin{equation}
\operatorname{ChannelAttention}(\mathcal{Q}_c, \mathcal{K}_c, \mathcal{V}_c) = \operatorname{softmax}\left(\frac{\mathcal{Q}_c \mathcal{K}_c^T}{\sqrt{d_k}}\right) \mathcal{V}_c
\end{equation}
where $\mathcal{Q}_c$, $\mathcal{K}_c$, and $\mathcal{V}_c$ represent the query, key, and value matrices for the channel dimension.

In the second stage, Time Attention aggregates tokens across different time segments within the same channel to capture the full temporal semantics of the input:
\begin{equation}
\operatorname{TimeAttention}(\mathcal{Q}_t, \mathcal{K}_t, \mathcal{V}_t) = \operatorname{softmax}\left(\frac{\mathcal{Q}_t \mathcal{K}_t^T}{\sqrt{d_k}}\right) \mathcal{V}_t
\end{equation}
where $\mathcal{Q}_t$, $\mathcal{K}_t$, and $\mathcal{V}_t$ are the query, key, and value matrices for the time dimension.

This two-stage attention mechanism is particularly powerful as it allows the model to handle data from any combination of sensor channels efficiently, adapting to different configurations with minimal computational overhead. The reduction in complexity from $O(N^2 \times M^2)$ to $O(N^2) + O(M^2)$ is crucial in preventing computational overload, thereby enabling real-time diagnostics and prognostics across various equipment types. Moreover, by focusing attention on the most relevant parts of the input data, this mechanism significantly enhances the model's ability to extract and utilize critical features, leading to more accurate and reliable diagnosis and prognosis outcomes.

The structure of the model is illustrated in Fig. \ref{fig:framework}b. RmGPT is composed of a patch-based tokenizer and $K$ stacked layers of the RmGPT Layer, which store the learnable prompt tokens, task tokens, and fault tokens.

\subsection{Training Process}

The training process of the RmGPT model involves two main stages: pretraining and downstream task adaption. Each stage plays a crucial role in ensuring that the model can generalize well to various diagnosis and prognosis tasks.

\subsubsection{Next Signal Token Prediction Pretraining}

We propose a pretraining strategy specifically designed for PHM applications, inspired by techniques from the natural language processing domain. The objective is to predict the next signal token in a sequence, which prevents the model from simply repeating patterns and forces it to learn the underlying dynamics of the system. Accurately predicting the next token indicates that the model has learned the intrinsic signal dynamics, which are crucial for representing the system health state. 

{%
For the RmGPT model, the input is a patch-tokenized multivariate sequence $x_p \in \mathbb{R}^{l_s \times M \times P}$, where each of the $M$ channels has been segmented into $l_s$ temporal patches of length $P$. During pretraining, we mask the final patch $x_{p_{l_s}}$ of each channel and train the model to reconstruct it based on the preceding patches $\{x_{p_1}, x_{p_2}, \ldots, x_{p_{l_s-1}}\}$.
}

{
Formally, the pretraining objective is to minimize the reconstruction loss between the true final patch and the predicted output from the model:
\begin{equation}
\mathcal{L}_{\text{pretrain}} = \left\| x_{p_{l_s}} - G(z_{l_s}) \right\|^2
\end{equation}
where $z_{l_s} \in \mathbb{R}^{M \times d}$ denotes the predicted latent vector corresponding to the masked token position, and $G(\cdot)$ is a lightweight decoder that projects this latent vector back to the patch space $\mathbb{R}^{M \times P}$.
}

{
The latent vector $z_{l_s}$ is computed by applying the Transformer encoder $F(\cdot; \theta_e)$ over the visible patch tokens:
\begin{equation}
[z_1, z_2, \ldots, z_{l_s}] = F(x_{p_1}, x_{p_2}, \ldots, \texttt{[MASK]}; \theta_e)
\end{equation}
and extracting the output at the masked position $z_{l_s}$. In other words, the model is trained to infer the representation at the masked location from the surrounding context, and then reconstruct the raw signal patch through the decoder head.
}

{
This formulation encourages the model to learn informative temporal representations by reconstructing raw signal content from the masked token embedding, rather than directly generating the signal. In our implementation, we use a fixed-length context of $n = 7$ patches to predict the 8th patch for each channel. This self-supervised design enables the model to capture localized dynamics and cross-channel dependencies in a semantically grounded and computationally efficient manner.
}

To adapt the pretrained RmGPT model to downstream diagnosis and prognosis tasks, we consider two complementary strategies: full finetuning and prompt learning.

\subsubsection{Prompt Learning}

To efficiently adapt the pretrained model to specific diagnosis or prognosis tasks, we adopt a lightweight prompt tuning framework, which introduces a small number of learnable task-specific tokens into the model input while keeping the pretrained backbone parameters $\theta_e$ frozen. Specifically, we incorporate three types of tokens:

\begin{itemize}
    \item \textit{Prompt Tokens} $T_p$: encode task- or equipment-specific information, guiding the model’s attention toward relevant signal features;
    \item \textit{Task Tokens} $T_t$: act as aggregation centers for multichannel signal representations, producing a global health descriptor;
    \item \textit{Fault Tokens} $T_f$: serve as semantic anchors for different fault types, enabling interpretable and similarity-based classification.
\end{itemize}

During training, only the parameters associated with these tokens—denoted as $\theta_p \subset \theta_e$—are updated, while the remaining pretrained parameters remain unchanged. This leads to a parameter-efficient adaptation strategy, requiring the tuning of less than 5\% of the total model parameters.

The prompt-enhanced input is constructed as $(T_s + T_p)$, and the training objective is to optimize only the prompt-related parameters $\theta_p$ based on task-specific supervision:

\begin{equation}
\theta_p = \arg \min_{\theta_p \subset \theta_e} \mathbb{E}_{(\mathcal{X}_i, \mathcal{Y}_i) \sim D} \left[ \mathcal{L}_{\text{task}}(\mathcal{Y}_i, F(\mathcal{X}_i + T_p; \theta_p)) \right]
\end{equation}

This design enables flexible adaptation in scenarios with limited training data or constrained computational resources, while reducing the risk of overfitting. The task tokens $T_t$ play a central role in summarizing multichannel signal features into a global health state, which is compared against fault tokens $T_f$ to produce interpretable and discriminative predictions.

\subsubsection{Full-Parameter Finetuning}

As an alternative to prompt learning, RmGPT also supports traditional finetuning, where all model parameters $\theta_f$ are updated using labeled data from the target task:

\begin{equation}
\theta_f = \arg \min_{\theta_f} \mathbb{E}_{(\mathcal{X}_i, \mathcal{Y}_i) \sim D} \left[ \mathcal{L}(\mathcal{Y}_i, F(\mathcal{X}_i; \theta_f)) \right]
\end{equation}

Here, $\theta_f$ shares the same dimensionality and structure as the pretrained parameters $\theta_e$, and can be viewed as a continuously optimized version of $\theta_e$ tailored to the target task. In other words, full-parameter finetuning initializes from $\theta_e$ and updates the entire parameter set to better fit the new data distribution.

This approach enables deeper task-specific adaptation and potentially stronger performance when sufficient labeled data is available. However, it also introduces higher computational overhead and a greater risk of overfitting, particularly in low-resource settings.
\subsubsection{Adaptation Methods: Prompt vs. Finetuning}
In contrast, prompt learning provides an efficient and lightweight alternative by updating only a small number of learnable task-specific tokens ($\theta_p$: $T_p$, $T_t$, $T_f$), while keeping the backbone frozen. This allows for fast adaptation across different tasks or domains, with lower risk of overfitting and reduced computational cost.

Table~\ref{tab:adaptation_comparison} summarizes the differences between these two strategies across key aspects, including parameter efficiency, expressiveness, and suitability for different scenarios.

\renewcommand{\arraystretch}{1.2}
\begin{table}[ht]
\centering
\caption{Comparison of Prompt Learning and Full Finetuning in RmGPT}
\label{tab:adaptation_comparison}
\begin{tabular}{@{}p{2cm}p{2.7 cm}p{2.7 cm}@{}}
\toprule
\textbf{Aspect} & \textbf{Prompt Learning} & \textbf{Full Finetuning} \\
\midrule
\textbf{Updated Parameters} & Only task-specific tokens ($\theta_p$: $T_p$, $T_t$, $T_f$) & All model parameters ($\theta_f$) \\
\textbf{Parameter Ratio} & Less than 5\% & 100\% \\
\textbf{Efficiency} & Fast, low computation & Slow, high computation \\
\textbf{Overfitting Risk} & Low & High\\
\textbf{Expressiveness} & Moderate (leveraging pretrained features) & High (full model adaptation) \\
\textbf{Use Cases} & Few-shot learning, cross-domain transfer & Full adaptation with sufficient labeled data \\
\bottomrule
\end{tabular}
\end{table}

Overall, the two strategies offer flexible trade-offs: prompt learning is more suitable for rapid deployment in resource-constrained environments, while full finetuning provides greater expressiveness and task-specific capacity when sufficient data and compute are available.

The overall training process first equips the model with general signal understanding through self-supervised pretraining, followed by lightweight or full adaptation to downstream tasks. Prompt learning enables efficient task transfer by updating only a small set of task-specific embeddings, while full finetuning allows complete adaptation when sufficient data is available. Together, these strategies make RmGPT a flexible and powerful foundation model for rotating machinery health management across varying equipment, signal modalities, and fault conditions.

\section{EXPERIMENTS}
In this section, we conduct a comprehensive evaluation of the RmGPT model across various diagnosis and prognosis tasks using diverse datasets. We begin by describing the datasets employed in our experiments, highlighting their diversity and relevance to real-world applications. Following this, we present the main results, comparing the performance of RmGPT to SOTA algorithms\cite{chang2023intelligent,sun2023liteformer,chen2021bearing,cui2024triplet,qin2024inverse,xu2021fault,chen2022dual}. Additionally, we assess the few-shot learning capabilities of RmGPT, demonstrating its effectiveness in scenarios with limited training data. Subsequently, we conduct an efficiency analysis of RmGPT during the training and inference processes. Finally, we analyze the token space representations learned by RmGPT, illustrating its ability to transform noisy input signals into a clear and distinct semantic space.

\subsection{Dataset description}

In this study, we employ a variety of datasets to evaluate the performance of the RmGPT model on both diagnosis and prognosis tasks. These datasets, summarized in Table \ref{tab:dataset}, originate from different types of equipment and exhibit a wide range of characteristics, including signal channels, fault categories, sampling frequencies and dataset sizes. 
\begin{table}[htbp]
\caption{{Overview of datasets used for diagnosis and prognosis tasks}}
\label{tab:dataset}
\centering
\setlength{\tabcolsep}{1pt} 
\renewcommand{\arraystretch}{1.2} 
\begin{tabular}{@{}lcccccc@{}}
\toprule
Tasks      & Object       & Datasets & Classes & Channels & Points & Frequency \\ \midrule
\multirow{5}{*}{Diagnosis} 
           & Bearing      & CWRU\cite{shi2024cross}& 16      & 1, 2, 3  & 9,011,145    & 12 kHz           \\
           & Bearing      & SLIET\cite{SLIET}& 13      & 3        & 13,631,475   & 70 kHz           \\
           & Gear         & QPZZ-II\cite{QPZZ}& 5       & 9        & 1,064,960    & 5120 Hz          \\
           & Gear         & SMU\cite{SMU}& 3       & 1        & 300,000      & 10 kHz           \\ \midrule
Prognosis  & Bearing      & XJTU\cite{XJTU}& 4       & 2        & 204,046,336  & 25.6 kHz         \\ \bottomrule
\end{tabular}
\end{table}

The above rolling bearing and gear datasets are popular benchmark datasets that have been widely utilized to validate the effectiveness of various algorithms. However, due to the significant differences between these datasets, no previous work has attempted to use a model with the same parameters across all seven diverse datasets for both diagnosis and prognosis tasks. This study seeks to demonstrate the feasibility of constructing a unified model with consistent parameters for fault diagnosis and prognosis in rotating machinery, using these varied datasets as a preliminary validation.

\subsection{Experimental Setup}
All experiments are conducted on a computing server equipped with 8 NVIDIA RTX 3090 GPUs. To ensure consistency across diverse datasets, we uniformly downsample all signals to approximately 5kHz and standardize input windows to 2048 time steps. This preprocessing harmonizes variations in signal amplitude and sampling rates, enabling the model to focus on learning meaningful temporal patterns.

{We adopt a unified training pipeline for all datasets. The training consists of two phases: a self-supervised pretraining stage and a supervised adaptation learning stage. In the first phase, the RmGPT model is pretrained for 20 epochs on the unlabeled training data aggregated from all datasets, enabling the model to learn generalizable temporal representations. In the second phase, we perform adaptation learning using labeled data, with two different adaptation strategies: prompt learning and full-parameter finetuning. The only difference between the two lies in the scope of trainable parameters—prompt learning updates only a small subset of task-specific tokens ($\theta_p \subset \theta_e$), whereas finetuning updates the entire parameter set ($\theta_f$) initialized from $\theta_e$. All other training configurations, including learning rate, optimizer, number of epochs, and batch size, are kept consistent across both adaptation strategies to ensure fair comparison.}

{The model architecture consists of 4 Transformer layers with 512 hidden units, totaling 68.50M parameters. This compact yet expressive configuration is selected to balance representation capacity and computational efficiency. The tokenizer uses a stride and patch length of 256, and the prompt and fault token lengths are set to 10 and 1, respectively.}

Table~\ref{tab:hyperparameters} summarizes the hyperparameter settings used throughout all experiments.

\begin{table}[htbp]
\caption{Hyperparameters Used for Training RmGPT Model}
\label{tab:hyperparameters}
\centering
\setlength{\tabcolsep}{2pt}
\renewcommand{\arraystretch}{1.2}
\begin{tabular}{@{}cc@{}}
\toprule
\multicolumn{2}{c}{\textbf{Hyperparameters}} \\ \midrule
Batch Size                      & 256 \\
Learning Rate                   & \(3.00 \times 10^{-7}\) \\
Pretraining Epochs              & 20 \\
Finetuning Epochs               & 3 \\
Prompt Learning Epochs          & 5 \\
Tokenizer Stride (S)            & 256 \\
Tokenizer Patch Length (P)      & 256 \\
Transformer Layers              & 4 \\
Hidden Size ($d$)               & 512 \\
Prompt Token Length ($l_p$)     & 10 \\
Fault Token Length ($l_t$)      & 1 \\
Total Model Parameters          & 68.50M \\
\bottomrule
\end{tabular}
\end{table}

\subsection{Main results}

We conducted a comprehensive evaluation of RmGPT across a wide range of diagnosis and prognosis tasks using multiple public rotating machinery datasets. To ensure fairness, all experiments adopt a consistent training-to-testing split ratio of 8:2. Following self-supervised pretraining on the aggregated unlabeled data, RmGPT is adapted to downstream tasks via two adaptation strategies: prompt learning and full-parameter finetuning. Except for the number of trainable parameters, all experimental settings are kept identical across these two adaptation modes. In contrast, all baseline models are trained individually on each dataset in a supervised manner, without benefiting from shared pretraining or a unified adaptation mechanism.

This rigorous and consistent evaluation framework highlights the robustness and versatility of RmGPT across heterogeneous PHM scenarios. A key advantage of our approach lies in the ability to use a single model configuration—with shared pretrained parameters and unified adaptation settings—across all datasets and tasks. As summarized in Table~\ref{tab:performance}, RmGPT consistently outperforms existing state-of-the-art (SOTA) methods, demonstrating its effectiveness as a general-purpose foundation model for rotating machinery health management.

In fault diagnosis tasks, both RmGPT-Prompt and RmGPT-Finetune configurations show strong performance. The prompt-based version achieves an average accuracy of 99.21\% across datasets, with individual accuracies ranging from 98.05\% to 100\%, while the finetuned version further improves to an average of 99.75\%, achieving perfect scores on several challenging datasets such as QPZZ and SLIET.

In prognosis tasks, RmGPT also delivers superior results. For RUL prediction on rolling bearing datasets, the RmGPT-Prompt configuration achieves a Mean Absolute Error (MAE) of 0.150 and a Mean Squared Error (MSE) of 0.033. The RmGPT-Finetune configuration achieves even better performance, reducing the errors to 0.136 MAE and 0.030 MSE.

These results collectively demonstrate the feasibility and advantage of constructing a unified PHM model that can generalize across diverse equipment types, signal characteristics, and task objectives. The ability to deploy a single model with consistent architecture and training strategy across heterogeneous datasets marks an important step toward scalable, foundation-level modeling in industrial prognostics and health management.

\begin{table*}[!htbp]
\caption{RmGPT Uses the Same Parameter Model and Outperforms SOTA Algorithms on Different Tasks and Equipments}
\label{tab:performance}
\centering
\setlength{\tabcolsep}{12pt} 
\renewcommand{\arraystretch}{1.2} 
\begin{tabular}{@{}ccccccccc@{}}
\toprule
Task            & \multicolumn{6}{c}{Diagnosis}                                                                                                  & \multicolumn{2}{c}{Prognosis}   \\ \midrule
Dataset         & CWRU               & QPZZ               & SLIET             & SMU                & \multicolumn{2}{c}{\multirow{2}{*}{Average\_ACC↑}}& \multicolumn{2}{c}{XJTU}        \\
Metric          & Accuracy↑          & Accuracy↑          & Accuracy↑         & Accuracy↑          & \multicolumn{2}{c}{}& MAE↓           & MSE↓           \\ \midrule
RmGPT-Prompt    & \underline{99.92\%}& \textbf{100.00\%}  & \underline{98.05\%}& \underline{99.11\%}& \multicolumn{2}{c}{\underline{99.21\%}}& \underline{0.150} & \underline{0.033} \\
RmGPT-Finetune  & \textbf{100.00\%}  & \textbf{100.00\%}  & \textbf{99.02\%}  & \textbf{100.00\%}  & \multicolumn{2}{c}{\textbf{99.75\%}}& \textbf{0.136}    & \textbf{0.030} \\
Transformer\cite{sun2023liteformer} & 99.39\%            & \textbf{100.00\%} & 91.02\%            & 65.67\%            & \multicolumn{2}{c}{89.02\%}& 0.231          & 0.070          \\

CNN\_LSTM\cite{chen2021bearing}         & 86.73\%            & 96.98\%           & 93.85\%            & 66.67\%            & \multicolumn{2}{c}{86.05\%}& 0.186          & 0.045          \\
DP\_MRTN\cite{chen2022dual}         & \textbf{100.00\%}  & \textbf{100.00\%} & 98.33\%            & 98.02\%            & \multicolumn{2}{c}{99.09\%}& 0.250          & 0.083          \\
IMSFACNN\cite{xu2021fault}          & 98.62\%            & 75.00\%           & 98.11\%            & 66.67\%            & \multicolumn{2}{c}{84.60\%}& 0.250          & 0.083          \\
ResNet18\cite{chang2023intelligent} & 99.71\%            & \underline{98.03\%}& 97.66\%           & 48.89\%            & \multicolumn{2}{c}{86.07\%}& 0.259          & 0.091          \\
InversePINN\cite{qin2024inverse}    & 99.48\%            & \textbf{100.00\%} & 96.29\%            & 80.67\%            & \multicolumn{2}{c}{94.11\%}& 0.159          & 0.035          \\
TARTDN\cite{cui2024triplet}         & 91.07\%            & 95.00\%           & 97.94\%            & 63.56\%            & \multicolumn{2}{c}{86.89\%}& 0.158          & 0.034          \\ \bottomrule
\end{tabular}
\end{table*}


\subsection{Few-Shot Learning Capabilities  }
To further evaluate the adaptability and robustness of the RmGPT model, we conducted few-shot learning experiments on the CWRU dataset. In these experiments, the training samples were limited to a few examples per class. Specifically, 1-shot learning indicates that there is one sample per class for all 16 classes, resulting in a total of 16 training samples. Similarly, 4-shot learning means there are four samples per class, totaling 64 training samples, and so on. The training set was used without any data augmentation. The number of testing samples was kept constant at 13,734. Importantly, all methods, including RmGPT and the compared models, were evaluated without any prior exposure to additional samples from the CWRU dataset, ensuring that none of the models had seen the test data during training. Moreover, the training and testing sample sizes were kept consistent across all models to guarantee a fair comparison. This rigorous approach ensures that the observed performance differences genuinely reflect the models' capabilities in few-shot learning scenarios. {%
 It is also important to note that while the CWRU dataset may be partially included in the unlabeled corpus used during the self-supervised pretraining stage, no label information was provided to the model at that time. The pretraining process relied solely on unlabeled signal sequences, without access to class identities or diagnostic labels. Therefore, the few-shot learning performance reflects the model’s ability to transfer general temporal representations learned from unlabeled data, rather than any memorization of labeled examples. This distinction ensures the validity of the few-shot learning evaluation and highlights the benefit of the foundation model’s generalization capability.
}The results of the few-shot learning experiments are summarized in Table~\ref{tab:fewshot}.

RmGPT consistently outperforms other models, demonstrating its ability to achieve high accuracy with very few training samples. In the 1-shot scenario, RmGPT-Prompt achieves a diagnosis accuracy of 82.46\% across 16 fault classes, significantly outperforming other models. In the 4-shot scenario, RmGPT-Prompt reaches an accuracy of 98.53\%, further highlighting its robustness and effectiveness. These results demonstrate that RmGPT can efficiently leverage small amounts of data to achieve high-precision diagnostics. RmGPT's exceptional performance in few-shot learning scenarios makes it highly suitable for practical equipment maintenance situations, where fault samples are often scarce. 

These results underscore the strength of RmGPT in practical scenarios, demonstrating its potential to be used as a universal model for various rotating machinery diagnostics and prognostics, even with limited fault data. The model's exceptional performance in few-shot learning scenarios can be attributed to several factors. First, RmGPT benefits from extensive pretraining on large datasets, which enables it to learn generalizable fault signal patterns. This pretraining helps the model extrapolate from minimal examples, making it highly effective in adapting to new fault categories with limited labeled samples. Second, the generative nature of RmGPT allows it to infer unseen fault patterns by leveraging its learned representations of signal semantics, a critical capability in few-shot settings where data is scarce. Lastly, RmGPT’s prompt learning mechanism facilitates rapid adaptation to new tasks by efficiently utilizing the few available training samples. This combination of pretraining, generative modeling, and efficient task adaptation makes RmGPT highly robust in few-shot learning scenarios, where limited data availability is often a challenge.

\begin{table}[htbp]
\caption{{Few-shot Diagnosis Performance on CWRU Dataset}}
\label{tab:fewshot}
\centering
\resizebox{0.95 \linewidth}{!}{%
\setlength{\tabcolsep}{0.5pt} 
\renewcommand{\arraystretch}{1.2} 
\small 
\begin{tabular}{@{}ccccccc@{}}
\toprule
\multicolumn{1}{c}{Training Data} & \multicolumn{6}{c}{Accuracy↑} \\ \cmidrule(lr){2-7}
Setting& 1-shot         & 4-shot         & 8-shot         & 16-shot        & 10\% Data& Full Data\\ \midrule
RmGPT-Prompt         & \textbf{82.46\%} & \textbf{98.53\%} & \textbf{99.27\%} & \textbf{99.549\%} & \textbf{100.00\%} & \textbf{100.00\%}\\
RmGPT-Finetune       & \underline{65.56\%} & \underline{68.84\%} & \underline{72.83\%} & \underline{86.86\%} & \underline{99.88\%} & \textbf{100.00\%} \\
Transformer          & 10.75\%        & 15.57\%        & 15.10\%        & 19.75\%        & 95.41\%        & \underline{99.39\%}\\
CNN\_LSTM            & 10.73\%        & 5.54\%         & 10.75\%        & 10.73\%        & 81.91\%        & 86.73\%        \\
DP\_MRTN             & 5.36\%         & 21.47\%        & 41.57\%        & 55.36\%        & 99.82\%        & \textbf{100.00\%} \\
IMSFACNN             & 5.42\%         & 5.42\%         & 5.42\%         & 5.42\%         & 28.18\%        & 98.62\%        \\
ResNet18             & 5.42\%         & 5.29\%         & 6.06\%         & 5.42\%         & 46.29\%        & 99.71\%        \\
InversePINN          & 5.37\%         & 10.78\%        & 14.28\%        & 5.37\%         & 95.08\%        & 99.48\%        \\
TARTDN               & 5.40\%         & 5.40\%         & 5.40\%         & 5.40\%         & 33.65\%        & 91.07\%        \\ \bottomrule
\end{tabular}
\normalsize 
}
\end{table}

\subsection{Efficiency Analysis}
To evaluate the efficiency of the RmGPT model, we compared its performance under various configurations, focusing on inference time, FLOPS, backward time, and the number of training parameters. The results are summarized in Table \ref{tab:efficiency}.

The impact of the proposed patch-based tokenization and channel-time Attention mechanism on reducing computational complexity and improving performance is significant. The original configuration shows an inference time of 4.970 ms and FLOPS of 51.097 GFLOPS. Removing the patch-based tokenization increases the inference time to 337.773 ms and FLOPS to 3813.998 GFLOPS. Similarly, excluding the channel-time Attention mechanism results in an inference time of 9.281 ms and FLOPS of 102.214 GFLOPS. For training efficiency, the prompt learning approach is more efficient than finetuning. The backward time for prompt learning is 6.765 ms with 3.038 M training parameters, compared to 22.023 ms with 68.548 M parameters for finetuning. 

\begin{table}[htbp]
\caption{{Efficiency Comparison of RmGPT}}
\label{tab:efficiency}
\resizebox{0.95\linewidth}{!}{%
\centering
\setlength{\tabcolsep}{2pt} 
\renewcommand{\arraystretch}{1.2} 
\begin{tabular}{cccc}
\toprule
\multicolumn{4}{c}{\textbf{Model Efficiency}} \\ \midrule
 & \textbf{Orin} & \textbf{Without Patch} & \textbf{Without TC Attention} \\ \midrule
\textbf{inference Time (ms)} & 4.970 & 337.773 & 9.281 \\ 
\textbf{FLOPS (GFLOPS)} & 51.097 & 3813.998 & 102.214\\ \midrule
\multicolumn{4}{c}{\textbf{Training Efficiency}} \\ \midrule
 & \textbf{Prompt} & \textbf{Finetune} &  \\ \midrule
\textbf{Backward Time (ms)} & 6.765 & 22.023 &  \\ 
\textbf{Training Parameters (M)} & 3.038 & 68.548 &  \\ \bottomrule
\end{tabular}
}
\end{table}

The model ability to process diverse signal inputs efficiently and adapt to various tasks using prompt learning demonstrates its potential as a versatile tool for rotating machinery diagnostics and prognostics.

\subsection{Token Space Analysis}
To further analyze the inference process of RmGPT, we visualize the original signal samples along with the corresponding Time-Frequency Task Tokens (CLS Tokens) and Fault Tokens obtained by feeding the signals into RmGPT. We applied the t-SNE method to reduce the dimensionality of these tokens to a 2D space for visualization.
\begin{figure}[!htbp]
\centerline{\includegraphics[width=1\columnwidth]{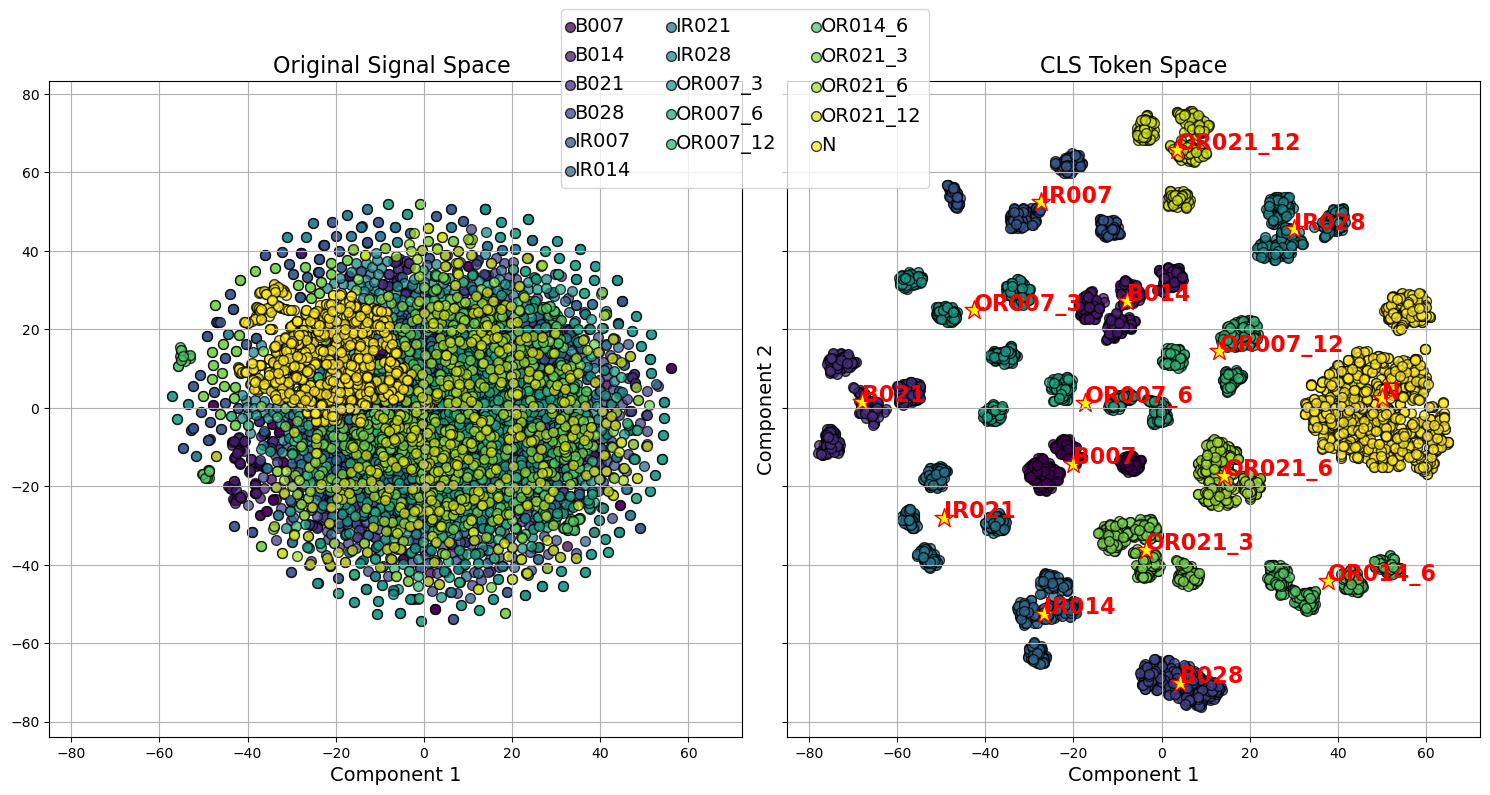}}
\caption{RmGPT can transform noisy input signals into a CLS semantic space with clear class boundaries}
\label{fig:t-sne}
\end{figure}

Fig. \ref{fig:t-sne} shows the comparison between the original input signal space and the CLS Token space generated by RmGPT. Different categories of samples are represented by different colors. The left Figure shows the dimensionality reduction of the original signals, whereas the right Figure shows the dimensionality reduction of the CLS Tokens, with the positions of the Fault Tokens for different categories marked in red. The visualized CLS Token space shows the much clearer classification boundaries than the original signal space. Additionally, the model Fault Tokens learn the prototype characteristics of different categories. Different fault modes are tightly clustered around distinct class centers, indicating that RmGPT accurately learns and distinguishes the intrinsic fault patterns in the data. This clear separation of fault classes in the CLS Token space highlights the model ability to transform noisy input signals into a more structured and interpretable representation, facilitating more accurate diagnosis and prognosis.

\section{Conclusion}

{This paper presents \textbf{RmGPT}, a foundation model for unified diagnosis and prognosis in rotating machinery. By utilizing a single model with consistent parameters across diverse datasets, RmGPT significantly outperforms state-of-the-art (SOTA) algorithms, demonstrating exceptional accuracy and robustness in both diagnosis and prognosis tasks. The model's ability to effectively handle varied equipment types, signal channels, fault categories, and sampling frequencies underscores its adaptability and validates the feasibility of a unified approach in the PHM domain.}

{The results of few-shot learning experiments further highlight RmGPT’s capability to generalize from limited data, achieving remarkable performance even with minimal training samples. The token space analysis reveals that RmGPT can transform noisy input signals into a structured semantic space with clear class boundaries, indicating its proficiency in learning and distinguishing intrinsic fault patterns.}

{\textbf{This work represents a pioneering attempt to explore the feasibility of foundation models in the field of rotating machinery.} RmGPT provides preliminary evidence that such models, which have shown great success in natural language and vision domains, can also be developed for complex industrial time-series scenarios. Our study not only validates the conceptual soundness of building a unified PHM foundation model, but also offers a practical framework that achieves superior performance across varied fault types and datasets.}

{Looking ahead, several important challenges and directions remain to be explored. From the perspective of \textit{depth}, future research may investigate the relationship between model scale and downstream performance within the rotating machinery domain, and evaluate whether large-scale foundation models like RmGPT can effectively address the highly challenging problem of cross-device generalization. From the perspective of \textit{breadth}, it would be promising to extend the generative token-based paradigm to other industrial time-series tasks—such as energy systems, batteries, wind turbines, or robotic health monitoring—thereby inspiring broader industrial applications of PHM foundation models.}

{Furthermore, we envision the development of a standardized capability evaluation framework for industrial foundation models, which would allow researchers to more systematically assess the robustness, adaptability, and scalability of such models under diverse industrial conditions. This would provide essential infrastructure for the next generation of intelligent and generalizable diagnostic systems in industry.}

\section{Acknowledgements}
This work was partially supported by Grant of National Natural Science Foundation (Grant No. 62371297) and Fund of Shanghai Engineering Research Center of Civil Aircraft Health Monitoring (Grant No. GCZX202204). The work of Enrico Zio is supported by iRel40 European co-funded innovation project, granted by the ECSEL Joint Undertaking (JU) under grant agreement No 876659.

\bibliographystyle{IEEEtran}

\bibliography{Reference.bib}


 




\vfill

\end{document}